**Article Title:** Clash of the models: Comparing performance of BERT-based variants for generic news frame detection

**Corresponding and first author:** Vihang Jumle; vihang.jumle@unibe.ch; ORCiD: 0009-0008-3388-9404; Address: Fabrikstrasse 8, 3012 Bern.

**Word count:** 6,932 words (including footnotes, references and tables)

*Declarations:*

**Ethical approval and consent to participate:** The design of the project, of which the article is part, has been approved by the Ethics Committee of the Faculty of Business, Economics and Social Sciences of the University of Bern (serial number 382023). No human participation was needed for this specific study.

**Consent for publication:** This article contains original work of the author and does not contain any copyrighted material.

**Availability of data and material:** The link to the models is provided through an anonymous link in the manuscript footnote; The models will be made available through a stable DOI post-publication.

**Competing Interests:** The author reports there are no competing interests to declare.

**Funding:** The author discloses receipt of the following financial support for the research, authorship, and/or publication of this article: This article is funded by the Swiss National Science Foundation [105217_215021].

**Authors' contribution:** This is a solo study, and the first author is credited with all the work.

**Acknowledgements:** This research is part of the Swiss National Science Foundation funded project "Algorithm audit of the impact of user- and system-side factors on web search bias in the context of federal popular votes in Switzerland" (PI: Mykola Makhortykh; ID: 105217_215021). The author thanks Linus Kiener for his help with data coding. The author also thanks Mykola Makhortykh, Maryna Sydorova, Vedant Jumle and Akriti Upreti for their feedback on various points of the draft.

**Note:** This is a pre-peer review version of the manuscript generated for archiving.

*Main Text*

**Abstract:** Framing continues to remain one of the most extensively applied theories in political communication. Developments in computation, particularly with the introduction of transformer architecture and more so with large language models (LLMs), have naturally prompted scholars to explore various novel computational approaches, especially for deductive frame detection, in recent years. While many studies have shown that different transformer models outperform their preceding models that use bag-of-words features, the debate continues to evolve regarding how these models compare with each other on classification tasks. By placing itself at this juncture, this study makes three key contributions: First, it comparatively performs generic news frame detection and compares the performance of five BERT-based variants (BERT, RoBERTa, DeBERTa, DistilBERT and ALBERT) to add to the debate on best practices around employing computational text analysis for political communication studies. Second, it introduces various fine-tuned models capable of robustly performing generic news frame detection. Third, building upon numerous previous studies that work with US-centric data, this study provides the scholarly community with a labelled generic news frames dataset based on the Swiss electoral context that aids in testing the contextual robustness of these computational approaches to framing analysis.

## Introduction

Framing continues to be one of the most extensively used communication theories amongst scholars interested in studying various discourses in society (Olsson & Ihlen, 2018). Proposed by Entman (1993), this concept conceptualises the making of certain aspects of social reality more prominent than others in one's mode of communication. The theory's versatility has enabled its application across sub-disciplines, including the study of communication on climate change (Vu et al., 2021), racism (Lane et al., 2020), and conflict (Tschirky & Makhortykh, 2024), among many others.

At the core of framing research, scholars attempt to identify the different 'frames' (defined aptly by de Vreese [2005: 53] as "a frame is an emphasis in salience of different aspects of a topic") that a certain form of communication material contains. Exposure to different frames, such as the economic implications of climate change versus the moral implications of climate change, is assumed to create differing perceptions amongst audiences of the same underlying reality, i.e. in this example, climate change. Identifying these frames, especially the salient ones and their reach either selectively (Messing & Westwood, 2014) or incidentally (Lee & Kim, 2017), therefore, becomes instrumental to explaining how public opinion comes to be shaped.

The recent decade has seen growth in the application of computational methods towards frame detection [1,2]. Although computational methods contrasted with the qualitative approaches that scholars had traditionally used to study frames, they enabled the detection of frames at scale. These methods include the machine learning models that use bag-of-word features, and more reliably the transformer models (Card et al, 2015; Liu et al 2019; Walter & Ophir, 2019; Ali & Hassan, 2022; Jumle et al, 2025; Burscher et al, 2014; Naderi & Hirst, 2017)[3]. Such growth and diversity in methods have naturally called for inter-method comparisons to identify the best performers and for an informed discussion on their strengths and weaknesses, as well as computational cost and best practices for performance assessment and validation. Studies and reviews suggest that transformer models often outperform models that use bag-of-words features (Khanehzar et al 2019; Ali & Hassan 2022; Author, n.d.), however, it is not clear so far how the different transformer models compare against each other. This is important because different models come with their own sets of advantages (such as improved classification) and disadvantages (higher computational costs or longer processing times). Being aware of these intricacies is essential for the scholarly community to make informed choices regarding their frame detection tasks in relation to available resources.

---

[1] Frame detection refers to the process of detecting pre-defined frames deductively within a corpus. This differs from frame identification which refers to the process of identifying newer frames inductively within a corpus. Topic modelling, for example, is a common method used for inductively identifying frames.

[2] In practice, the methodology of frame 'detection' and frame 'classification' can imply the same process. A model can be said to be first trained on a labelled dataset and then *classify* these learned frames present within an unseen

This article addresses this gap by presenting a comparison of frame detection performance among five key transformer models frequently used for classification tasks: BERT, RoBERTa, DeBERTa, DistilBERT, and ALBERT. Specifically, it conducts generic news frame detection (Semetko and Valkenburg (2000) – a prominent typology used to compare journalistic coverage - on a corpus of search engine results produced against search queries on two popular votes related to pension payments (from 2024) and the environment (from 2025) in Switzerland, adding to the existing work that tends to focus more on US cases.

---

corpus. This frame classification is synonymous with frame detection, where the model detects the learned frames within an unseen corpus.

[3] Computational frame analysis runs its own set of conceptual limitations in capturing frames as compared to interpretive analysis. Refer to Ali & Hassan (2022) for a discussion on these limitations.

The rest of the paper is organised as follows. First, the article briefly engages with related work on transformers and their use for automated frame detection. This is followed by a description of transformer models and how they are employed by this article towards frame detection (such as the model architecture, hyperparameters and validation). The significance of the used dataset is also explained here. The comparative results are presented toward the end, followed by a discussion on the models' performance, and the ways the debate on best practices for using computational text analysis for frame detection may be extended further by the scholarly community.

**Related Work**

This study engages with the literature on computational frame analysis, particularly on two fronts: on the efficacy of transformer models (and the gap in intra-transformer comparisons) and commonly used datasets.

First, studies and reviews have suggested that transformer models (like BERT and its variants) outperform older models that use bag-of-words features when used for automated frame detection (Ali & Hassan, 2022). Their attention mechanisms enable the contextual analysis of texts, allowing frames to be detected more effectively than previous approaches (Kuang et al., 2025). Some studies present a clear winning case, such as Galgoczy et al. (2022), where BERT outperformed Decision Trees, Naïve Bayes, Logistic Regression, and Support Vector Machines in sentiment frame classification. Likewise, Mendelson et al. (2021) present RoBERTa's better performance compared to Random Predictor and Logistic Regression in the classification of immigration frames. Other studies reveal a mixed scenario, where not all, but some, transformer models outperform these previous methods. For instance, consider the cases of Reveilhac &

Morselli (2022) and Sanchez-Junquera et al. (2021), where transformer models outperformed older approaches in some cases, such as when training datasets were small and limited (Stede et al., 2023). While the consensus is not absolute, studies generally suggest that transformer models are likely better by a notable margin compared to models that use out-of-bag features.

The adoption of transformer models, however, faces roadblocks since the current peer-reviewed literature on intra-transformer model performance for frame classification remains limited. There are many transformer models, and each model performs differently in the training context (i.e., the dataset theme, number of classes, hyperparameter settings, and architectural setup). It is imperative, therefore, to conduct systematic comparisons of a broad range of curated models to enable their informed use by the wider scholarly community for frame classification.

Existing studies that provide such crucial comparisons fall short on two fronts: on the one hand, these studies experiment with only a few models, commonly BERT, its improved version RoBERTa, and BERT's distilled version, DistilBERT. Some key examples include Alonso del Barrio and Gatica-Perez (2023), Khanehzar et al. (2021), Lyu & Takikawa (2022), and Lin et al. (2022). Excluding other models, such as DeBERTa, which is proposed as an improvement over RoBERTa, or ALBERT (a lite BERT variant), leaves a gap in the literature and limits potential improvements in automated frame detection. On the other hand, these comparisons (or even single model assessments) are rarely conducted systematically. Studies tend to engage with hyperparameter tuning in a limited manner. Many studies limit themselves to furnishing their best model's configuration. Only a few studies explain how they arrived at the best configuration and detail their search algorithm (Carosia, Coelho & Silva, 2020; Eisele et al, 2023). Not having these details keeps the scholarship shortsighted regarding the efficacy of different models in various training contexts. Additionally, rarely do studies provide metrics or perspectives on costs of computation compared to their performance, leaving the discussion on the affordability of automated frame detection (which is a key research design consideration) largely unaddressed (Izsak et al, 2021).

Second, datasets play a crucial role in determining the performance of various transformer models. The various text complexities, sentence structures, and socio-economic themes contained in the text all strongly influence how transformers perform any form of classification, including frame classification or detection. Literature has worked with various datasets, many of which apply these models to social media data (Galgoczy et al., 2022; Tyagi et al., 2022), news articles (Guo et al., 2023), or political speeches (Bonikowski et al., 2022). Search engine data remains rare, which this study contributes to the scholarship. What makes search engines a unique source of political information is that they bring users to a curated bundle of mixed sources, from short-form material like tweets to long-form material like party manifestos, in addition to other sources like blogs, advertisements, and websites of institutions, which have yet to see an application of computational framing using transformers. Curating such a dataset for a non-US case helps diversify the current scope of framing research.

Data and Methods

*Working with generic news frames*
The paper compares the performance of various transformer models for detecting generic news frames. The idea of generic news frames comes from Semetko and Valkenburg (2000). They group generic news frames as those that commonly feature in journalistic reportage. Their persistence across cases makes them a key working concept for comparative communication research. Generic news frames consist of five types (the brackets suggest an abbreviation that the paper uses): Attribution of Responsibility (AR01), Human Interest (HI02), Conflict (CF03), Morality (MF04) and Economic Frame (EF05). The paper defines frames as Semetko and Valkenburg (2000); however, it introduces a few operational variations. While Semetko and Valkenburg (2000) used an extended set of questions per frame to assess frames' presence in a corpus, this paper instead condenses these extended questions down to up to two per frame. The questions were drafted carefully such that the frames' core meaning was captured. This condensation was a logistical necessity as it simplified manual labelling of the training dataset.

The paper examines the detection of six frames, including five generic news frames and an additional 'No Frame' (NO06), which are defined below. A paragraph was used as the unit of coding, as coding at the sentence level was insufficient to capture individual frames, whereas complete documents consisted of multiple frames. Paragraphs sometimes contained multiple frames; however, the study only coded for the dominant frame through a close reading of each paragraph.

The author trained a coder to conduct the coding based on the codebook. The author and the coder then coded a trial set, discussed disagreements, until a reliable intercoder agreement was achieved.

*Attribution of Responsibility (AR01)*

1. Does the paragraph suggest that a specific individual (e.g. a politician) or a group (e.g. a party, state, governmental departments, society, civilian groups) is responsible for the problem or can resolve it?
2. Does the paragraph suggest a solution to the problem or call for urgent action over it?

*Human Interest Frame (HI02)*

1. Does the paragraph use a human example or emphasise the effect of a problem on humans? Is the human the central focus of the paragraph?

2. Does the paragraph use emotive language that may invoke an emotional response in the reader (like outrage, empathy-caring, sympathy, or compassion)?

*Conflict Frame (CF03)*

1. Does the paragraph refer to any form of negative interaction or framing (disagreement, confrontation, spat, etc.) between two sides of any kind (actors, problem, viewpoints)?

*Morality Frame (MF04)*

1. Does the paragraph contain any form of morality (what is good or bad)? Does it talk about Good and Evil?

2. Does it talk about religious tenets or prescribe socially apt behaviour or ethics?

*Economic Frame (EF05)*

1. Does the paragraph speak about economic changes in policy or law, or refer to any form of economic loss/gain, expense, costs, or economic consequences of a current policy or law?

*No Frame (NO06)*

The paragraph does not fit any frame-related criteria outlined above.

**Training Dataset**

The training data was collected from webpages via an algorithmic audit of two search engines (Google and Bing) in the early months of 2024 and 2025. These webpages were collected by systematically querying two search engines with a set of search queries. These search strings were created by surveying the Swiss population on how they would search for information about the Swiss popular votes on pension and retirement ages (henceforth, Pension Initiatives) in March 2024 and the environmental responsibility initiative (henceforth, Environment Initiative) in February 2025 (refer to Vziatysheva et al 2024 and Vziatysheva 2025 for the respective survey details). Designed as a virtual agent-based audit (Ulloa et al., 2024), the study deployed virtual agents to collect the top ten text search results (most of which were media sites, federal institution sites, and other voting-related outlets) and save them as HTML files. Once the data collection was completed, the study extracted the main text from these webpages and divided it into paragraphs as they appeared on the respective webpages. Paragraphs that were not in English were translated into English using the Google Translate API. Data was stored in a tabular format, with a unique webpage ID assigned to a set of paragraphs that came from the same webpage.

A dataset for the model was randomly extracted from this corpus, which, at the end, consisted of 2,959 paragraphs (1,887 paragraphs from webpages of Pension Initiative and 1,072 paragraphs from webpages of Environment Initiative). The aim was to create a training dataset that contained a few hundred rows per frame (like some studies that use four or more frames for classification) and attempted to achieve this with the available resources for manual coding. Care was taken to ensure no paragraphs were blank and did not contain irrelevant information, like cookie notices[4], and were long enough to meaningfully contain a frame (which the study operationalised by including only those rows that were a complete sentence at a minimum).

100 randomly selected paragraphs of these 2,959 paragraphs were labelled independently by the author and a trained coder. A satisfactory Cohen's Kappa was achieved. The coder then labelled all the 2,959 paragraphs, a distribution of which is provided below.

| Frames | Count (Percentage) |
| --- | --- |
| Attribution of Responsibility (AR01) | 226 (7.6%) |

---

[4] The study conducted keyword searches (for instance, "cookie") to extract such cookie notices.

| | |
|---|---|
| Human Interest (HI02) | 32 (1%) |
| Conflict (CF03) | 124 (4.2%) |
| Morality (MF04) | 14 (0.4%) |
| Economic (EF05) | 542 (18.3%) |
| No Frame (NO06) | 2,021 (68.3%) |
| **Grand Total** | 2,959 |

Table 01: Distribution of manually labelled generic news frames

The dataset structurally lacks paragraphs with frames. The study initially labelled a thousand paragraphs and found the overall frame count to be too low; hence, it was expanded twice to approximately three thousand paragraphs, which is when three of the five frames (AR01, CF03, EF05) attained an acceptable training volume. The study also attempted to review a few hundred extra paragraphs within the remaining unlabelled data, specifically for MF04; however, this search did not yield any results. In the interest of time and available research resources, the study proceeded with the set reported in Table 01.

**Empirical Testing**

*Model selection*

The sub-sections below describe the study's approach to preparing the data and initialising the architecture for the set of employed transformer models. Only the best configurations are described in these sub-sections. These were determined through exhaustive experimentation, which involved varying key hyperparameter levels.

The study employs these five models for specific reasons, such as aiding comparison and decision-making for communication scholars. BERT (Devlin et al. 2019; the study uses bert-base-uncased with 110M parameters) is naturally included because it serves as the basis for most transformer architectures and has been extensively used for frame detection (Lin et al, 2022). RoBERTa (Liu et al, 2019; the study uses roberta-base with 125M parameters) is picked as a case for a similar reason as BERT, since many studies have already used RoBERTa (Bonikowsi and Stuhler, 2022) as one of the models within their comparative frameworks, often standing out as the best performer (compared to machine learning models that use bag-of-word features and its immediate predecessor BERT). This extensive literature facilitates a comparison of the results of this study with existing performance benchmarks for RoBERTa.

On the contrary, DeBERTa (He et al, 2021; the study uses microsoft/deberta-v3-base with 184M parameters) is picked as a case because the literature has yet to test it actively for frame detection (as compared to BERT, RoBERTa, and the different variants) despite it being an improved version of RoBERTa. DeBERTa, therefore, presents itself as a unique model that the study aims to present as a benchmark for the scholarly community, comparing whether DeBERTa outperforms its relatively popular predecessor, RoBERTa, in generic news frame detection. DistilBERT (Sanh et al, 2020; the study uses distilbert-base-uncased with 66M parameters) is included as a case because it is a distilled version of BERT, which helps the study compare the performance trade-offs between computation-heavy and computation-light transformer models. DistilBERT, like BERT and RoBERTa, has been actively used by scholars for frame detection (Müller and Proksch, 2023), further making it easier to contrast against existing benchmarks. Lastly, ALBERT (Lan et al, 2020; the study uses albert-base-v2 with 12M parameters), however, is a model that has been severely overlooked by the scholarly community. ALBERT, with its factorisation of the embedding matrix, parameter sharing, and Sentence Order Prediction, has a smaller parameter size compared to BERT; hence, it is an apt comparative case for computation-heavy models and a potential alternative to DistilBERT. The study also aims to establish a performance benchmark for ALBERT.

*Data Preparation and Class Imbalance Management*

The set of 2,959 paragraphs was first processed to exclude NO06 and MF04. NO06 was excluded because the study focused solely on identifying a particular type of generic frame[5]. MF04 was excluded since relevant paragraphs (only 14) were significantly underrepresented as compared to all other classes, and including it within the model added noise and impacted the overall model performance. This produced a training corpus of 924 paragraphs.

These 924 paragraphs were stratified and split in a 4:1 proportion (or 80%-20%) to create a training set (739 paragraphs consisting of 180 AR01, 26 HI02, 99 CF03, and 434 EF05) and a test set (185 paragraphs). The test set was kept aside and underwent no further modifications, and was only used later for model validation. The train set was augmented to address the severe class imbalance. The study employed back-translation augmentation through Helsinki-NLP's OPUS-MT group of models (Tiedemann & Thottingal, 2020) to convert English paragraphs into German and French, and then back to English, thereby inflating all classes except EF05 to 250 paragraphs each (i.e., approximately half the size of the largest class). This produced an augmented training set of 1,184 paragraphs (AR01, HI02, CF03 each at 250 and EF05 at 434 paragraphs). The remaining class imbalance between EF05 and the rest of the frames was addressed by using inverse frequency weighting in the loss function. The study calculates class weights as [Weight = Total Samples / (Number of Classes x Count of Classes)], which assigns minority classes a higher weight and majority classes a lower weight. In practice, this approach adjusts the loss to penalise minority class errors more than the majority class. All frames except EF05 receive a weight of 1.18, whereas EF05 receives a weight of 0.68.

---

[5] It was also found experimentally that the models performed better when this classification was conducted as a two-step process. First, a classification of NO06 against all other Frames (i.e. a binary classifier) and then a multi-class classification to identify the different generic news frames.

*Model Architecture and Hyperparameters*

All models were architecturally configured in the same manner. All model architectures contain a pre-trained transformer backbone, followed by a pre-classifier layer (projecting a 768-dimensional vector to 512 dimensions), then by a ReLU activation and dropout regularisation (with a rate of 0.1), and finally a linear classification layer mapped to the four frames. All models are trained and tested on the same datasets, evaluated using the same metrics, and follow the same training procedure (such as warm-up steps, class weights, weight decay, and epochs). Since all models share a similar design at their core, we can assume that a common architectural configuration and training protocol do not disadvantage any model over the others. Differences in performance, therefore, can be reliably attributed to the respective BERT variants.

However, the different models, due to their specific designs, perform optimally for a different set of key hyperparameters, with learning rates or batch sizes being two of them (Izsak et al, 2021). For instance, practitioners observe that DeBERTa tends to be more sensitive to learning rates than BERT or RoBERTa, and that could affect its performance in certain training contexts. Likewise, distilled models like DistilBERT can work with higher batch sizes and learning rates as compared to the rest. This study, therefore, takes a customised approach to finding the best model performance metrics. The study conducts a grid search over varying learning rates and batch sizes for all the models to determine the best pair based on the highest Average Macro F1 score. This allows each model the opportunity to report its peak performance, hence attributing the performance differences to the respective model's data processing design.

The hyperparameters – up to six training epochs, weight decay of 0.01 (L2 regularisation), and Epsilon of 1e-8 (for AdamW optimiser) – remain constant across all models. The learning rates of 2e-5, 3e-5, 4e-5, and 5e-5, along with batch sizes of 8, 16, and 32, were tested for all models. DeBERTa's evaluation also included a learning rate of 1e-5 in addition to the aforementioned range to account for its sensitivity to lower learning rates. Each model was trained separately and evaluated after every epoch. The model that produced the highest Average Macro F1 score across the epochs was retained. To enable stable learning, the study implemented warm-up steps set to (the standard) 10 % of the total steps. This is especially helpful for BERT-based models that tend to have sensitive gradients. The learning rate scheduler was set up with Cosine annealing. The study also enabled the use of Automatic Mixed Precision and Gradient Clipping (max norm = 1.0), which helped balance speed and stability during training and reduce memory usage. The next section presents figures and metric tables on model performance[6].

## Performance Assessment

Validation of the models was carried out on the test set that contained a total of 185 paragraphs (Support for the four frames is: AR01 = 46, HI02 = 06, CF03 = 25, EF05 = 108). This section is organised around four figures and two tables, each of which conveys key insights on model performances for generic news frame detection.

First, the heatmap below in Figure 01 presents a top-view of the best Average Macro F1 (henceforth, F1) for each model over the range of learning rates and batch sizes. This heatmap suggests that though all these models share a similar basic architecture, their performance over

---

[6] The fine-tuned models and test data can be anonymously accessed at this link for testing: https://osf.io/z69me/overview?view_only=81aacf77118d43338b83eec2759d15da

Page 15 of 30

the same classification task and the same set of hyperparameters varies substantially. Inter-model performance difference (best score versus the worst score, on a scale of 0 to 1) is 0.15. Intra-model performance for BERT (best performer) is 0.11. The average difference between the best and worst scores for all models is approximately 0.1.

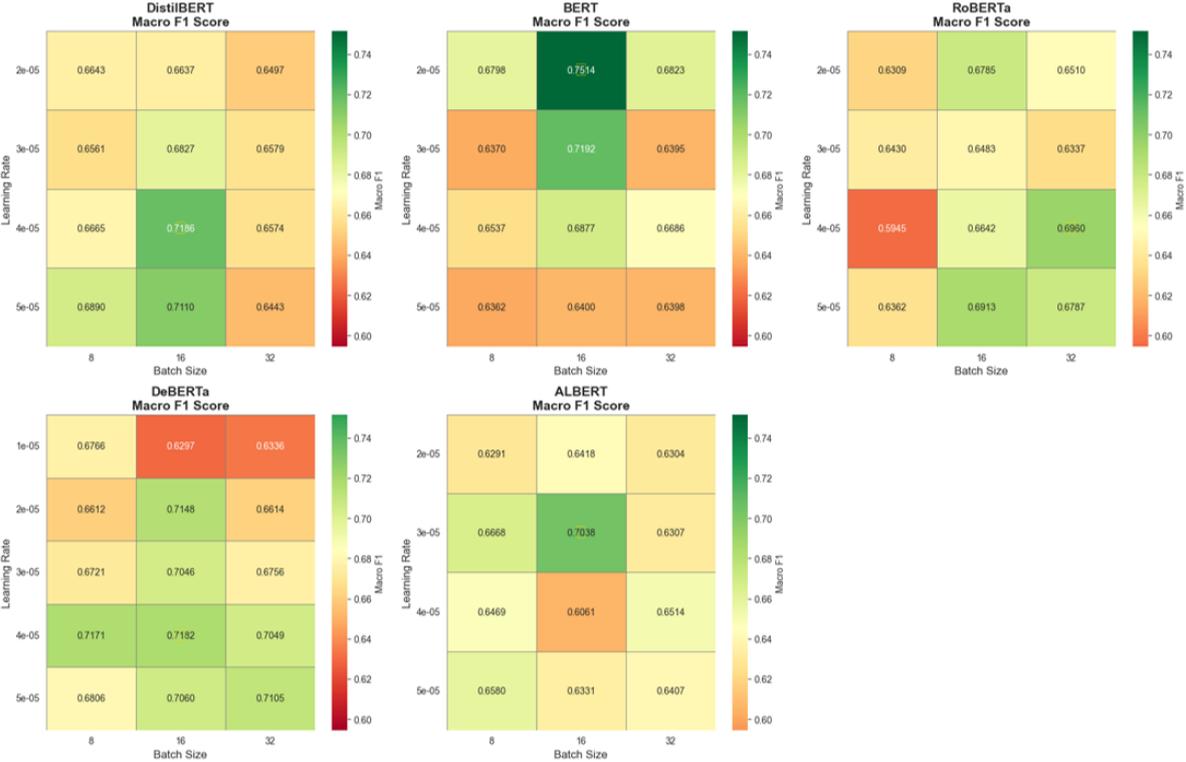

Figure 01: Heatmap of model performance over the two hyperparameter ranges. Each heatmap corresponds to a model and describes its respective scores for hyperparameter pairs.

We may visualise the heatmap data differently, as in Figure 02, to better observe the peculiarities of model performances. Here, we can observe that each model shows varying sensitivity to the learning rates. At some constant batch sizes, the trend in performance may remain similar (such as BERT at batch sizes 8 and 32); however, such trends cannot be generalised from the observed performance data. On the contrary, all models (except ALBERT) and, in most cases (DeBERTa at the lowest learning rate being an exception), exhibit a peak in performance at a batch size of 16 for all tested learning rates. For ALBERT, this holds true clearly only for one learning rate (3e-05) and to a limited extent for 2e-05.

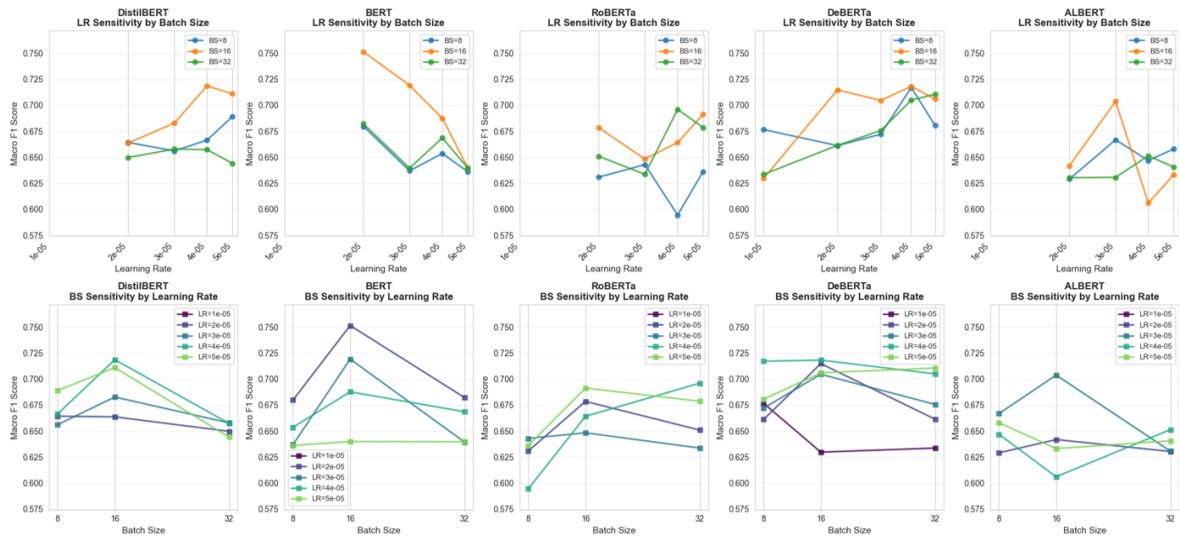

Figure 02: Line charts of model performance over hyperparameter ranges

Second, visualising the performance scores as a box plot helps us observe how the models tend to perform on average over the range of hyperparameters. A key insight (and that can be read in relation to the heatmap above) concerns the comparison of BERT to DeBERTa. We observe that the heaviest variant, DeBERTa, although it is not the best performer, produces a better average performance compared to all other models. Whereas BERT, which is the best performer, tends to perform relatively poorly in more than half of the cases. BERT's best performance is an exception produced for a specific hyperparameter configuration rather than a norm, which is contrary to DeBERTa, which produces stable, acceptable scores over the range of tested hyperparameters.

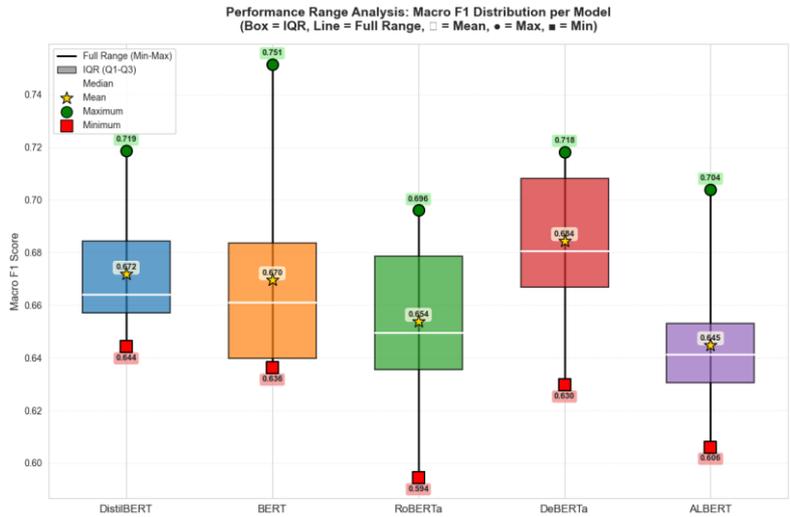

Figure 03: Box plot of model performance. Each box represents a model.

Third, Table 02 and Figure 04 below provide evaluation metrics at the class level for the best-performing model (details of the best-performing model are provided in Table 02). This allows us to observe how the meanings of different frames likely affect models' classification abilities.

|  | AR01 (46) | | | HI02 (6) | | | CF03 (25) | | | EF05 (108) | | |
|---|---|---|---|---|---|---|---|---|---|---|---|---|
| Model | P | R | F1 | P | R | F1 | P | R | F1 | P | R | F1 |
| DistilBERT | 0.58 | 0.65 | 0.61 | 1.00 | 0.50 | 0.67 | 0.77 | 0.80 | 0.78 | 0.83 | 0.80 | 0.81 |
| BERT | 0.67 | 0.63 | 0.65 | 0.80 | 0.67 | 0.72 | 0.71 | 0.88 | 0.76 | 0.85 | 0.83 | 0.84 |
| RoBERTa | 0.73 | 0.52 | 0.60 | 0.60 | 0.50 | 0.55 | 0.70 | 0.92 | 0.79 | 0.82 | 0.86 | 0.84 |
| DeBERTa | 0.69 | 0.63 | 0.66 | 0.57 | 0.67 | 0.62 | 0.73 | 0.76 | 0.75 | 0.85 | 0.86 | 0.85 |
| ALBERT | 0.60 | 0.54 | 0.57 | 0.75 | 0.50 | 0.60 | 0.81 | 0.84 | 0.82 | 0.81 | 0.84 | 0.82 |

Table 02: Best performance metrics per class

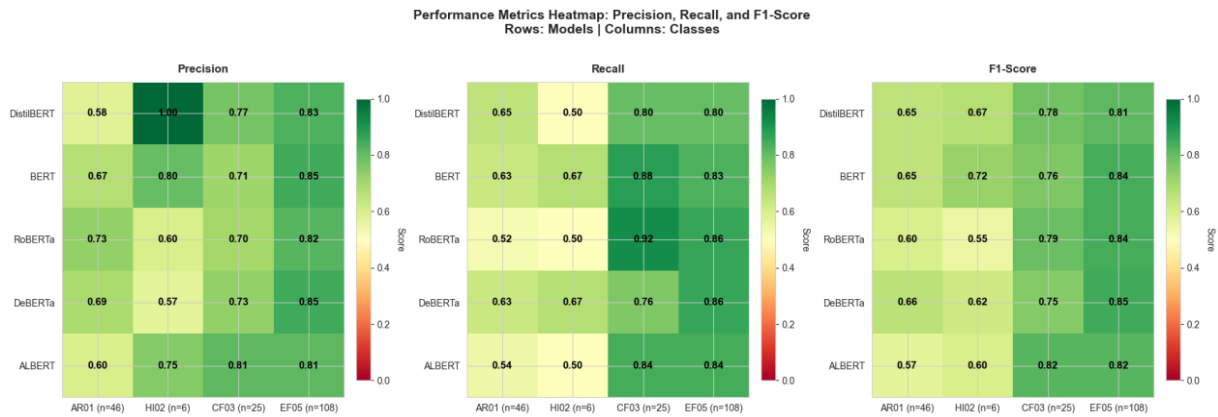

Figure 04: Heatmap of evaluation metrics per class

There are a few points to note here, and interpreting the metrics should be done keeping the context of the original class distribution in the training set in mind. Although the training data was augmented to balance the classes, it does not necessarily imply that the diversity of textual material (such as various examples and meanings) contained within each frame also reached a balance. In other words, although the models process all classes at the same volume, this does not imply that the quality of the classes was on par. Frames that are more prevalent in the original dataset (such as EF05 or AR01) can be assumed to be classified more accurately than those that are less prevalent. The size of the test set represents the proportion of classes that were present in the training set (since the data was stratified and split) and can assist in reading the metrics.

For both precision and recall, all models perform notably well for the high-volume class, like EF05. However, it remains difficult to generalise any further if the volume of classes directly affects the model performance. Consider CF03, which is lower in representation in the original set as compared to AR01, but most models perform better on its classification than AR01 (DeBERTa being the exception). On the other hand, there remain some model-specific peculiarities that likely extend an advantage or disadvantage to some models' performance in classifying a specific class over the rest. RoBERTa is a notable case that overall falls behind all models; however, it outperforms the rest significantly on recalling CF03. Likewise, DeBERTa, which closely follows BERT as the top performer, notably falls behind all models in recalling CF03, whereas ALBERT displays the highest precision for CF03.

| Model | Learning Rate | Batch Size | Macro F1 | Compute time in minutes (6 epochs) | Accuracy | Best N of Epoch |
|---|---|---|---|---|---|---|
| DistilBERT | 4e-5 | 16 | 0.72 | 2.40 | 0.75 | 6 |
| BERT | 2e-5 | 16 | 0.75 | 3.50 | 0.78 | 3 |
| RoBERTa | 4e-5 | 32 | 0.70 | 4.63 | 0.77 | 2 |
| DeBERTa | 4e-5 | 16 | 0.72 | 8.75 | 0.78 | 4 |
| ALBERT | 3e-5 | 16 | 0.70 | 5.79 | 0.76 | 4 |

Table 03: Best performance configuration per model

Discussion and Conclusion

The study centres its discussion and concluding remarks around a question: considering these model comparisons, how should the political communication scholarship approach automated frame detection, particularly for generic news frames?

First, Table 02 provides a unique guiding perspective. It suggests that, in certain situations as this, at an optimal hyperparameter level and the right training context, models – light or heavy – can all potentially produce acceptable performance. Here, all models reach an average Macro F1 of at least 0.7 (at their custom settings), which is near the acceptable reported floor for frame detection tasks (Author, n.d.). This is not to imply that all models can perform the same, but rather that in some training contexts, they may be able to do so. Certain performance differences persist, and it is worth noting that the path to determining the optimal hyperparameter is demanding. The study, therefore, invites the scholarly community to use these benchmark metrics as a reference to apply more datasets, models and varying hyperparameter settings, explore potential performance improvements and discuss the training contexts that lead most models to hit a similar performance ceiling.

Second, though different models contain the potential to produce robust performance, the path to reach those configurations may be practically costly. All the tested models produce an acceptable Macro F1 for a batch size of 16, which may not be feasible for interested scholars (let alone 32, as with RoBERTa) if they simply lack the computational architecture (such as GPUs) to train such models. It is, therefore, imperative for interested scholars to first assess the availability of computational resources and let that inform them of their choice of models and how they set their models' performance expectations. For instance, in cases where computation resources are limited, scholars are better off defaulting to DistilBERT or ALBERT with a lower batch size and accepting a slightly lower performance score compared to the heavier models, which may perform worse at lower batch sizes. Alternatively, scholars could adopt better memory management practices, such as gradient accumulation, that could help them simulate higher batch sizes than what their computational architecture would permit.

Third, scholars should also consider time constraints as part of assessing their resources. Besides computational bandwidth, time bandwidth also plays a role in helping models converge to their optimal peak performance. The provided computation time and epochs could serve as a benchmark to help scholars assess which models are better suited for systematic testing and which they might overlook when working under stricter time constraints. The models were computed on a Tesla T4 GPU, and the computation time is specific to this resource. Computing the same models on a CPU would take substantially longer. The DeBERTa grid search for this study was computed for 135 minutes, whereas BERT was computed for 40 minutes.

Fourth, the relevance of assessing computational and time resources is underscored when considering how often different models produce acceptable performance (refer to Figures 01 and 03). An apt comparison is between BERT and DeBERTa, the two top performers in our case. In case scholars would prefer to work with BERT, they must keep in mind that BERT performed exceptionally well only at one specific hyperparameter configuration and are recommended to invest time in systematically searching for the best hyperparameters. This could be both time- and computationally resource-intensive. On the other hand, if scholars prefer to work with DeBERTa, they could default to a batch size of 16 and compare the performance of different learning rates, making this model, though computational resource-intensive, potentially less demanding on time resources. Figure 02, likewise, suggests that if scholars need to make assumptions about hyperparameters, they may do so for batch sizes (say, working only with 16) and experimenting with learning rates, as the latter is shown to affect performance more than the former.

Lastly, scholars could consider two limitations of this study regarding computational frame detection. For one, this study was logistically constrained from ensuring absolute reproducibility in model performances. While the study sets random seeds, TRUE CuDNN deterministic, seeds for data loaders, weight initialisation, and data splits, it was constrained from ensuring deterministic CUDA operations and uses autocast() and GradScaler in Mixed Precision. Ensuring absolute reproducibility came at a substantial increase in resources, and the study accepted a slight numerical variation in evaluation metrics during a complete remodelling. Another shortcoming is that the use of grid search for optimal hyperparameters comes with its own methodological limitations, such as limiting the discovery of hyperparameter levels to local maxima, rather than the preferred global maxima. Refer to Alibrahim & Ludwig (2021) for a detailed discussion, and scholars are recommended to test other search approaches, such as Bayesian Optimisation or Genetic Algorithm. Comparative benchmarks generated through different search logics would substantially contribute to the literature on generic news frame detection and classification using transformers at large.

Lin, S.-Y., Kung, Y.-C., & Leu, F.-Y. (2022). Predictive intelligence in harmful news identification by BERT-based ensemble learning model with text sentiment analysis. Information Processing & Management, 59(2), 102872. https://doi.org/10.1016/j.ipm.2022.102872

Liu, S., Guo, L., Mays, K., Betke, M., & Wijaya, D. T. (2019). Detecting Frames in News Headlines and Its Application to Analyzing News Framing Trends Surrounding U.S. Gun Violence. In M. Bansal & A. Villavicencio (Eds.), Proceedings of the 23rd Conference on Computational Natural Language Learning (CoNLL) (pp. 504–514). Association for Computational Linguistics. https://doi.org/10.18653/v1/K19-1047

Liu, Y., Ott, M., Goyal, N., Du, J., Joshi, M., Chen, D., Levy, O., Lewis, M., Zettlemoyer, L., & Stoyanov, V. (2019). RoBERTa: A Robustly Optimized BERT Pretraining Approach (No. arXiv:1907.11692). arXiv. https://doi.org/10.48550/arXiv.1907.11692

Lyu, Z., & Takikawa, H. (2022). Media framing and expression of anti-China sentiment in COVID-19-related news discourse: An analysis using deep learning methods. Heliyon, 8(8). https://doi.org/10.1016/j.heliyon.2022.e10419

Mendelsohn, J., Budak, C., & Jurgens, D. (2021). Modeling Framing in Immigration Discourse on Social Media. Proceedings of the 2021 Conference of the North American Chapter of the Association for Computational Linguistics: Human Language Technologies, 2219–2263. https://doi.org/10.18653/v1/2021.naacl-main.179

Messing, S., & Westwood, S. J. (2014). Selective Exposure in the Age of Social Media: Endorsements Trump Partisan Source Affiliation When Selecting News Online. Communication Research, 41(8), 1042–1063. https://doi.org/10.1177/0093650212466406